%% file: main.tex
\documentclass[sigconf]{acmart}

\usepackage{microtype}
\usepackage{graphicx}
\usepackage{subfigure}
\usepackage{booktabs} 
\usepackage{multirow}

\usepackage{amsthm}
\usepackage{amssymb}
\usepackage{amsmath}
\usepackage{graphicx}
\usepackage[colorinlistoftodos]{todonotes}
\newcommand{\reals}{\mathbb{R}}

\renewcommand\footnotetextcopyrightpermission[1]{} 
\usepackage{hyperref}

\begin{document}

%
\title{Layered SGD: A Decentralized and Synchronous SGD Algorithm for Scalable Deep Neural Network Training}

%
\author{Kwangmin Yu}
\email{kyu@bnl.gov}
\orcid{0000-0003-0826-074X}
\affiliation{%
  \institution{Brookhaven National Laboratory}
  \streetaddress{P.O. Box 5000}
  \city{Upton}
  \state{New York}
  \postcode{11973-5000}
}

\author{Thomas Flynn}
\email{tflynn@bnl.gov}
\affiliation{%
  \institution{Brookhaven National Laboratory}
  \streetaddress{P.O. Box 5000}
  \city{Upton}
  \state{New York}
  \postcode{11973-5000}
}

\author{Shinjae Yoo}
\email{sjyoo@bnl.gov}
\affiliation{%
  \institution{Brookhaven National Laboratory}
  \streetaddress{P.O. Box 5000}
  \city{Upton}
  \state{New York}
  \postcode{11973-5000}
}

\author{Nicholas D'Imperio}
\email{dimperio@bnl.gov}
\affiliation{%
  \institution{Brookhaven National Laboratory}
  \streetaddress{P.O. Box 5000}
  \city{Upton}
  \state{New York}
  \postcode{11973-5000}
}

%
\renewcommand{\shortauthors}{Yu and Flynn, et al.}
\renewcommand{\shorttitle}{Layered SGD}

%
\begin{abstract}
Stochastic Gradient Descent (SGD) is the most popular algorithm for training deep neural networks (DNNs). As larger networks and datasets cause longer training times, training on distributed systems is common and distributed SGD variants, mainly asynchronous and synchronous SGD, are widely used. Asynchronous SGD is communication efficient but suffers from accuracy degradation due to delayed parameter updating. Synchronous SGD becomes communication intensive when the number of nodes increases regardless of its advantage. To address these issues, we introduce Layered SGD (LSGD), a new decentralized synchronous SGD algorithm. LSGD partitions computing resources into subgroups that each contain a communication layer (communicator) and a computation layer (worker). Each subgroup has centralized communication for parameter updates while communication between subgroups is handled by communicators. As a result, communication time is overlapped with I/O latency of workers. The efficiency of the algorithm is tested by training a deep network on the ImageNet classification task.

\end{abstract}

%
%


%

%
\maketitle

\input{introduction}

\input{related_work}

\input{distributed_SGD}

\input{algorithm}


\input{results}

\input{conclusion}


\bibliographystyle{elsarticle-num}
\bibliography{references}

\end{document}

%% file: introduction.tex
\section{Introduction}\label{sec:introduction}

\begin{figure*}
\centering
\includegraphics[width=\textwidth]{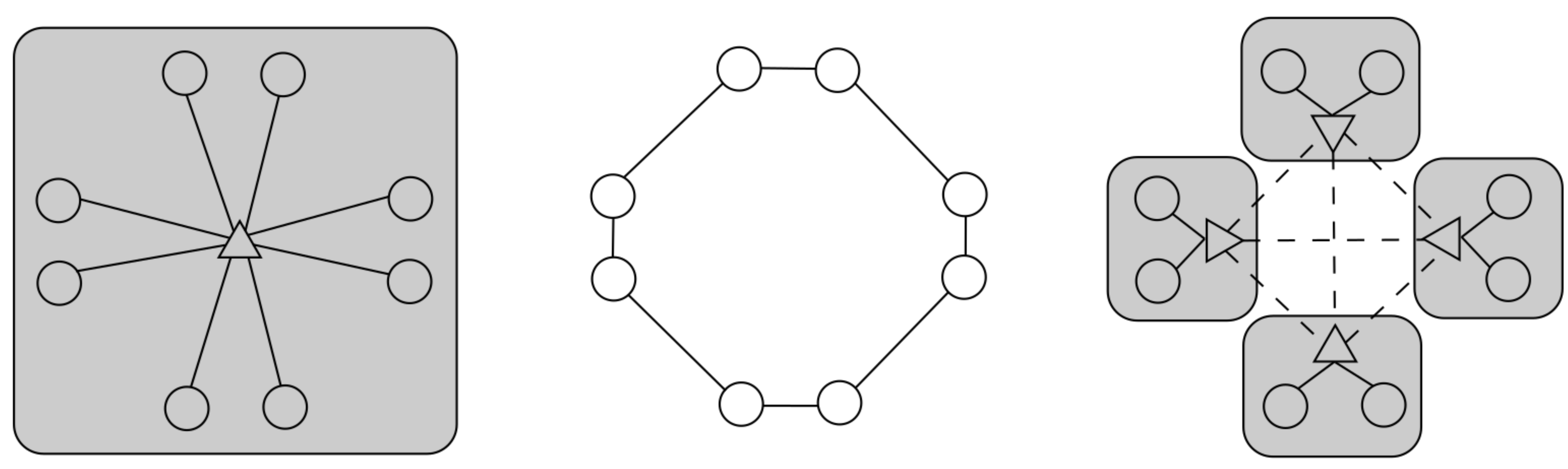}
\caption{Communication patterns in different distributed SGD algorithms. The left figure reflects the pattern in parallel SGD, where computation nodes synchronize via a central parameter server. The middle figures shows a possible topology in decentralized SGD, where computation nodes only synchronize with neighbor nodes. On the right is the communication pattern in Layered SGD, where local clusters, each containing one parameter server, are linked together.\label{fig:commpattern}}
\end{figure*}


Training machine learning models is typically framed as an optimization problem, where the objective is to minimize the error of the model measured relative to a database of training examples. In many cases the optimization is performed with gradient descent, and as dataset size increases and models become more complex (think large, deep neural networks), it becomes imperative to utilize not only accelerated devices within a compute node, such as multiple GPUs on a single machine, but to spread training across multiple physical nodes, resulting in a distributed system. This enables system designers to exploit hardware resources in order to reduce training time.

Moving SGD to multiple nodes raises the question of how should the underlying algorithm be adapted to make the most of the potentially vast computational resources available in a given distributed system. In SGD, the main operation that is repeated over and over again is the computation of gradients over a minibatch of examples. In this operation, the backpropagation procedure has to be carried out for each example and the resulting derivatives averaged. Since the derivative computations for different examples can be performed independently, this procedure lends itself to a simple data parallel approach: The gradient computation for a minibatch of size $M$ can be divided among $N$ workers, each of which compute $M/N$ derivatives, and after this computation the derivatives can be averaged together using an aggregation operation such as all-reduce. This algorithm, termed Parallel SGD, has demonstrated good performance, but it has also been observed to have diminishing returns as more nodes are added to the system. The issue is that the amount of time spent on the aggregation phase inevitably increases as the amount of nodes increases. Ideally, parallelization of SGD would yield linear scalability, but it appears that parallel SGD exhibits sublinear scalability. This is demonstrated as part of our experiments in Section \ref{sec:main_results}.

This begs the question of whether it is possible to find a variant of parallel SGD with reduced communication overhead, that still allows for fast optimization.  From an algorithmic standpoint, Parallel SGD is a centralized procedure: after each round of updates, all the nodes synchronize their parameters. Ensuring this consistency is the source of the communication bottle neck. Theoretically, the gradient computations are more useful if they are all performed with respect to the same set of parameters (that is, from the analysis standpoint, unbiased updates are better than biased ones.) However, the motivation behind decentralized algorithms is that perhaps we can get the most out of a distributed system by only requiring partial synchronization. Although there is some degree of error introduced in such procedures, this is countered by increased throughput; since less time is spent on communication, more epochs can be executed by the system in a given amount of time. Depending on the balance of these two factors, the decentralized approach may offer superior performance.


In this work we are motivated by the scenario where multiple fast and small local clusters are to be interconnected such as interconnected nodes equipped with multiple GPUs. It is assumed that nodes within a local cluster have low latency communication and that communication across clusters is more expensive. For this setting we introduce a new decentralized, synchronous, and hierarchical variant of SGD called Layered SGD (LSGD), that is described in detail in section \ref{sec:LSGD}. Briefly, the nodes are partitioned into groups, each group consisting of several computation nodes and one communication-only node that can be viewed as a parameter server. Communication between groups is managed by the communication-only nodes that serve as links not only between the nodes within a group, but between groups as well. While the nodes within a group are performing I/O (e.g., loading images onto GPUs), the group of communication nodes are performing an averaging operation among themselves. At the end of this operation, the local groups average their new weights with the communication node's new weights.
The main contribution of LSGD is that LSGD preserves the mathematical formula of the SGDs in Algorithm $1$ and $2$ (Algorithm $1$ and $2$ are same in the mathematical point of view, discussed in section \ref{sec:distributed_sgd}) in the sense that the parameters have exactly same values under the same conditions of data, hyperparameters, and initial parameters ($w_0$).
This is the main difference with other decentralized SGD algorithms \cite{decentralized, horovod, imagenet4min} which update parameters from its neighbour. 
Therefore, we claim LSGD has same accuracy with respect to the conventional SGD (Algorithm $1$ and $2$)
 
We summarize our main contributions as follows. 
\begin{enumerate}
\item We introduce the Layered SGD Algorithm, a new decentralized and synchronous approach to SGD that is adapted to distributed systems that consists of groups of high-performance clusters that are connected to one another via comparatively slower interconnects.
\item We carry out a set of detailed experiments of our algorithm that confirm the scalability properties of our algorithm, relative to conventional distributed SGD.
\end{enumerate}

%% file: related_work.tex
\section{Related work}
 We review work in two areas: Theoretical studies of stochastic gradient descent and its distributed variants, as well as systems-level papers on engineering deep learning systems.
 
 \paragraph{Stochastic Gradient Descent} Gradient descent and its variants have been studied for many years, and in recent years interest in these algorithms has been renewed due to their success on the optimization problems arising in machine learning. For standard (non-parallel) SGD, the pioneering work of \cite{ghadimi-lan} provided one of the first non-asymptotic analyses for SGD applied to non-convex functions, showing that the algorithm which returns a random iterate from the sequence generated by SGD is guaranteed (in a probabilistic sense) to be an approximate stationary point. For analogous results in the case of SGD for convex and strongly convex functions, see also \cite{bachmoulines, rakhlin} and the review in \cite{nemirovski}.
 
\paragraph{Parallel / Distributed SGD} An analysis of synchronous, parallel SGD for strongly convex function can be found in \cite{zinkevich}. For asynchronous parallel SGD, a non-asymptotic analysis for the case of convex functions is given in \cite{agarwalduchi}. An analysis of parallel and possibly asynchronous SGD for non-convex functions was carried out in \cite{lian2015}. 
In Elastic Averaging SGD \cite{elastic}, the authors took an approach in which the relation between the worker servers and parameter server is more flexible; workers synchronize with the parameter server only periodically, instead of every iteration, and when they do they only take a partial average of the parameter servers variables. Hence the workers do not share (or even try to share) the same parameters as in other parallel SGD algorithms.
  
Decentralized Parallel SGD (DPSGD) (for several classes of functions including non-convex ones) was considered in the recent work \cite{decentralized}. The primary difference is that in DPSGD all the nodes are treated the same, having both communication and computation functions. In our work, the nodes are organized in a hierarchical fashion; the "leaf" nodes are computation nodes and are connected to communication-only nodes that are responsible for communication across groups of leaf nodes. See Fig. \ref{fig:commpattern} for a comparison of several of the approaches.
 
\paragraph{Large Minibatch SGD} 
A separate strand of work is on engineering deep learning systems to achieve maximum throughput and the fastest training times. These focus on the ImageNet dataset and attempts to engineer the deep learning system to achieve state of the art results. These begin with \cite{alexnet, distbelief}. In \cite{imagenet1hr}, an approach to neural network training is presented that can achieve state-of-the art accuracy on ImageNet in less than one hour, by using a special learning rate schedule and overlaying communication with back propagation. Several authors have also noted the role of minibatch size in distributed training; large batch-sizes lead to increased throughput, but seem to negatively impact generalization. However, it was shown in \cite{imagenetin15} that several modifications to the training procedure, including a special learning rate schedule and a customized batch-normalization step, could improve the test accuracy in the large  minibatch regime, and in turn reduce the time to reach state of the art performance. 
Also, the relation between the learning rate and the minibatch size is studied in \cite{imagenet1hr} in detail.
Other modifications including the use of separate adaptive learning rates for each layer (LARS) \cite{You_CoRR2017, imagenetinminutes} and mixed-precision training \cite{imagenet4min} have been shown to bring down training times even further with high accuracy.
Even though our work is not directly related with increasing the accuracy of ImangeNet classification, we conduct tests based on ResNet-50 and ImageNet 2012. 
Since LSGD is an algorithm for SGD computation on distributed environment, it can be combined with accuracy related methods such as warmup learning rate \cite{imagenet1hr}, mixed-precision \cite{imagenet4min}, and LARS \cite{You_CoRR2017, imagenetinminutes}.
In our experiment, gradual warmup \cite{imagenet1hr} is adapted in both the CSGD and the LSGD implementations.

%% file: distributed_SGD.tex
\section{Distributed SGD}\label{sec:distributed_sgd}

Let us first set up the notation we will use.
We consider minimizing a function $f:\reals^n \to \reals$ that decomposes into a sum of $|X|$ functions from a set $X$:
\begin{equation}\label{eqn:optproblem}
\min_{w\in \reals^n} f(w) := \frac{1}{|X|}\sum\limits_{x \in X} l(w, x)
\end{equation}
The parameters $w$ could be, for instance, the weights in a neural network, and the loss function $l(w, x)$ represents the loss of our model on data $x \in X$ and parameter $w$. We denote by $\partial_w l (w, t)$ the derivative of the loss function $l (w,x)$ with respect to the parameter $w$. 


\begin{table}
\begin{center}
\label{alg:CSGD:single}
\begin{tabular}{rl}
    \hline
    \multicolumn{2}{l}{\textbf{Algorithm 1:} Conventional (non-distributed) SGD } \\ 
\hline \hline
    \multicolumn{2}{l}{\textbf{require}: 
    Initial parameter $w_0$,
    step size $\epsilon > 0$, } \\
    \multicolumn{2}{l}{number of iterations $T$} \vspace{0.5em}\\
\hline

1: & for all $t=0, \cdots, T-1$\\ 
2: & \hspace{5mm} Randomly draw mini-batch $M$ from $X$\\
3: & \hspace{5mm} Initialize $\Delta w = 0$\\
4: & \hspace{5mm} for all $x \in M$\\
5: & \hspace{1cm} aggregate update $\Delta w \leftarrow \partial_w l (w_t, x)$ \\
6: & \hspace{5mm} end for\\
7: & \hspace{5mm} update $w_{t+1} \leftarrow w_t - \epsilon \Delta w$\\
8: & end for \\ 
9: & return $w_T$ \vspace{0.5em}\\ 

\hline
\end{tabular}
\end{center}
\end{table}

\begin{table}
\begin{center}
\begin{tabular}{rl}

    \hline
    \multicolumn{2}{l}{\textbf{Algorithm 2:} Conventional distributed SGD with $N$ workers} \\ 
\hline \hline
    \multicolumn{2}{l}{\textbf{require}: 
    Worker index $i = 0, \cdots, N-1$, initial parameter $w_0$,}\\
    \multicolumn{2}{l}{
    step size $\epsilon > 0$, number of iterations $T$} 
    \vspace{0.5em}\\
\hline

1: & for all $t=0, \cdots, T-1$\\ 
2: & \hspace{5mm} Randomly draw mini-batch $M^i$ from $X$\\
3: & \hspace{5mm} Initialize $\Delta w^i = 0$\\
4: & \hspace{5mm} for all $x \in M$\\
5: & \hspace{1cm} aggregate update $\Delta w^i \leftarrow \partial_w l (w_t, x)$ \\
6: & \hspace{5mm} end for\\
7: & \hspace{5mm} \textbf{Allreduce $\Delta w^i$} over all workers and divide by $N$\\
8: & \hspace{5mm} update $w_{t+1} \leftarrow w_t - \epsilon \Delta w^i$\\
9: & end for \\ 
10: & return $w_T$ \vspace{0.5em}\\ 

\hline
\end{tabular}
\end{center}
\end{table}

The steps of conventional, non-parallel SGD and the conventional distributed SGD (CSGD) are summarized in Algorithm $1$ and Algorithm $2$, respectively. 
The routine from line $4$ to $6$ in both Algorithm $1$ and $2$ represents aggregating update of $\Delta w = \frac{1}{|M|} \sum\limits_{x \in M} l(w_t, x)$ and $\Delta w^i = \frac{1}{|M^i|} \sum\limits_{x \in M^i} l(w_t, x)$, respectively.
Usually, we have $\{ M_i \}$ is a partition (disjoint subset of $M$) of $M$ with same cardinality. Then we have $|M| = |M^i| N$.

The main difference in the two algorithms is the Allreduce operation and division by $N$ after aggregating $\Delta w$ by every sample in a mini-batch.
That is, Algorithm $2$ computes the average of all $\Delta w^i$.
This is the key feature of synchronous parallel SGD, which conserves all parameters, keeping them the same over all workers. 

Therefore, Algorithm $1$ and $2$ are same in the mathematical point of view.
When we have a mini-batch $M$ (for Algorithm $1$) and its partition $\{ M^i \}$ (for Algorithm $2$) where $i=0, \cdots , N-1$, both algorithms have same parameter values $w_t$ and $w_{t+1}$ before line number $4$ and after the line for $w_{t+1}$ update under the same conditions such as same hyperparameters, and same initial parameters ($w_0$).

\begin{figure}
\centering
\includegraphics[width=0.99\linewidth]{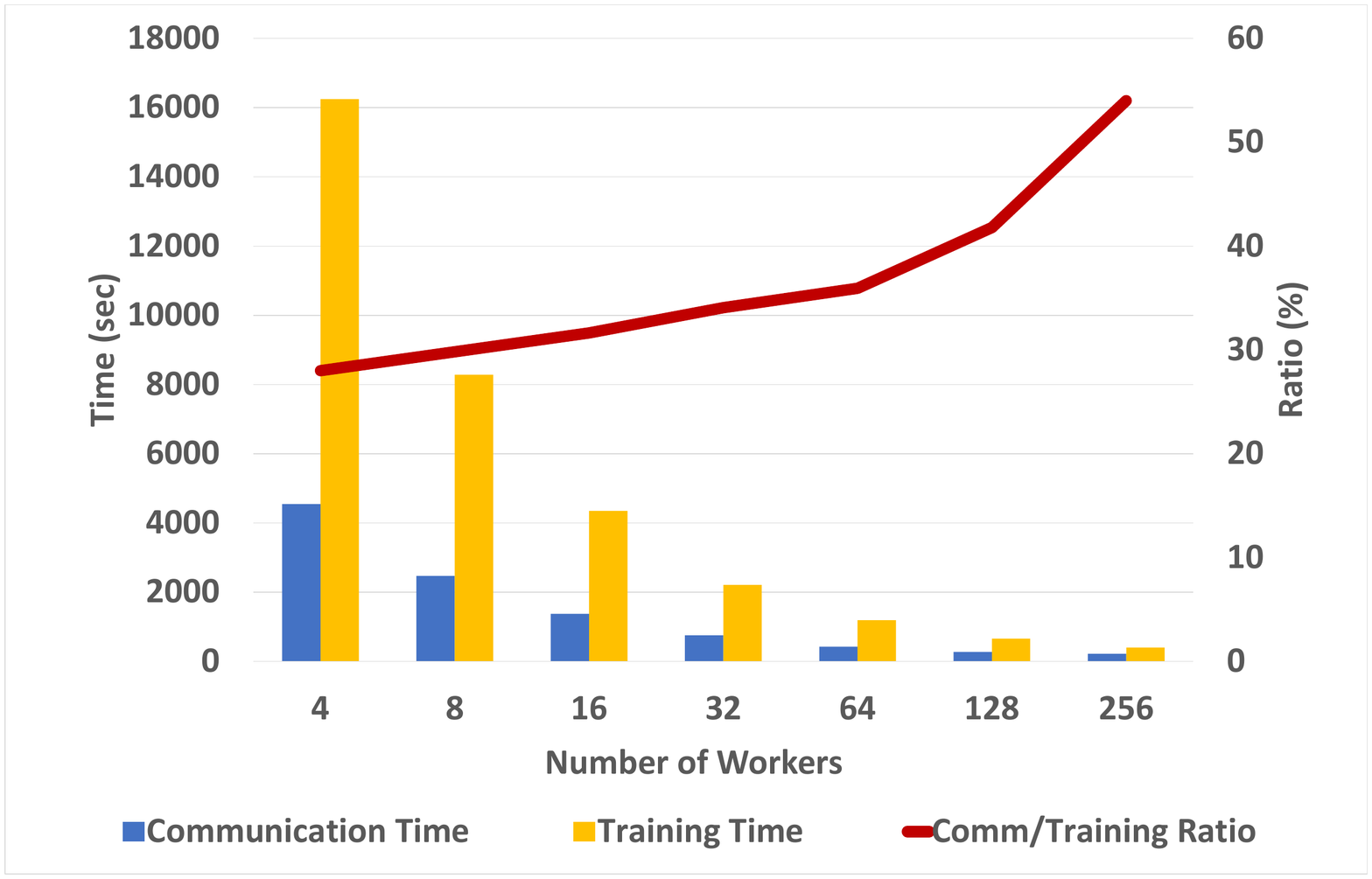}
\caption{Training time and Allreduce time (and their ratio) of conventional distributed SGD in an epoch (averaged). The local minibatch size is 64 per worker. The global minibatch size is 64 times the number of workers. Running environments are described in Section \ref{sec:main_results} in detail.}
\label{fig:CSGD}
\end{figure}

In spite of the advantage of accuracy, synchronous SGD suffers from an increasing communication burden as the number of  workers increases.
As shown in Fig. $\ref{fig:CSGD}$, the ratio of the Allreduce communication time to training time per epoch linearly increases after $64$ workers. Note that the total communication time decreases as the number of workers increases, as more workers increase the whole global minibatch size and thus requires fewer iterations to complete an epoch.
In fact, the increasing ratio of communication is the key reason of poor scalability of the CSGD beyond certain number of nodes (or workers).
To address this issue, in the next section we propose a new decentralized SGD algorithm that reduces the communication time by overlapping communication with I/O. 

%% file: algorithm.tex
\section{Layered SGD}
\label{sec:LSGD}

\begin{figure}
\centering
\includegraphics[width=\linewidth]{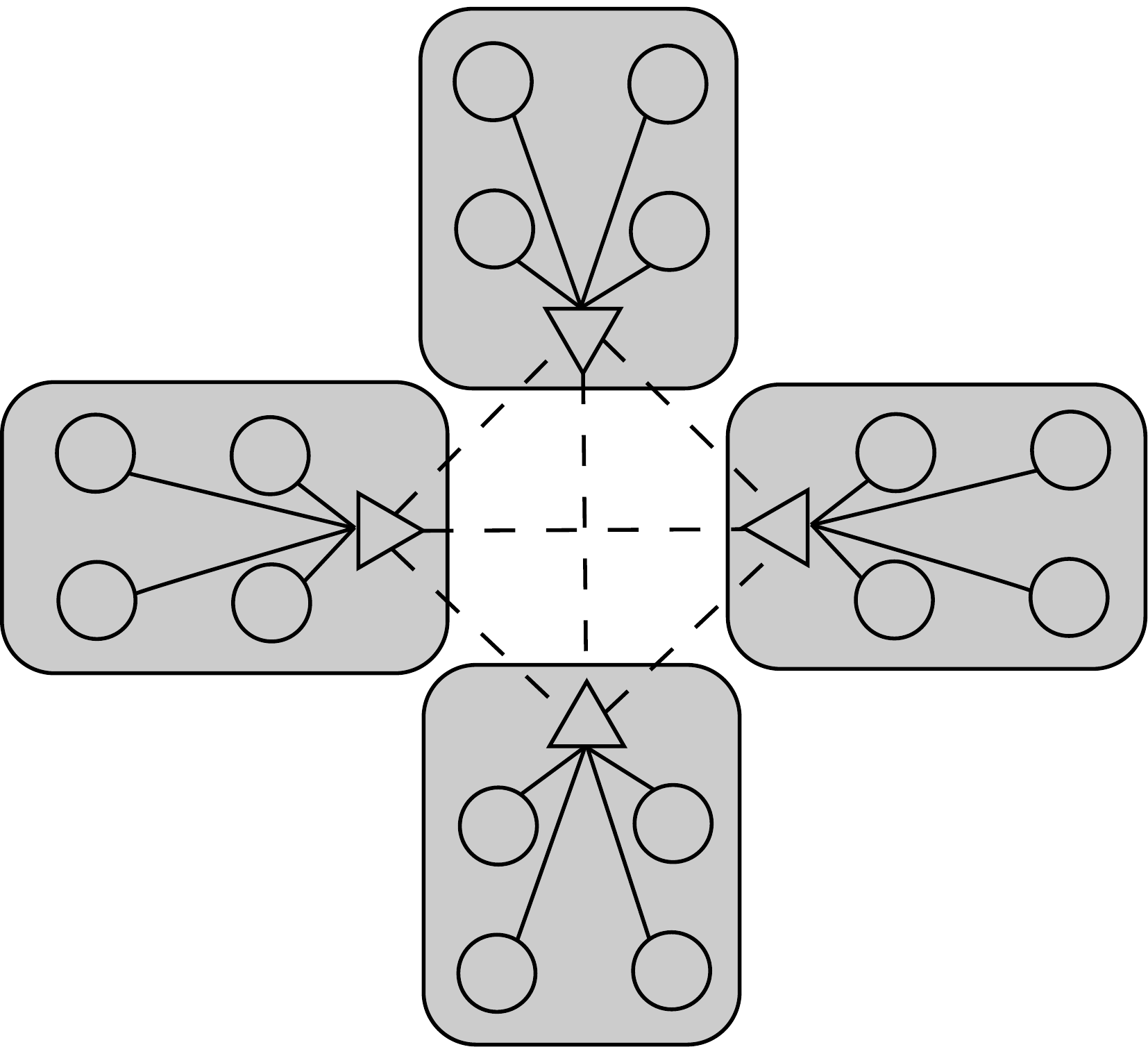}
\caption{Illustration of communication pattern in Layered SGD. Computation units (circles), typically GPUs, are grouped in a node (grey round rectangle), each containing a communicator (triangles), which plays a similar role as the local parameter server, take an Allreduce operation over all nodes. The solid edges represent links between computation and parameter servers. The dashed lines are links between parameter servers.}
\label{LSGD_diag}
\end{figure}

We propose a decentralized and synchronous SGD algorithm termed Layered SGD. 
We assume the system is accelerated by multiple GPUs.
Note that after describing the algorithm with this assumption, we can easily generalize to system consisting of only CPUs.

The main novelty of the algorithm is hiding communication burden of synchronous SGD as well as keeping the mathematics in Algorithm $1$ and $2$.

\begin{table*}
\begin{center}
\begin{tabular}{c|r|l|l}

    \hline
    \multicolumn{4}{l}{\textbf{Algorithm 3:} Layered SGD sequence} \\ 
\hline \hline
    \multicolumn{4}{l}{\textbf{require}: 
    Worker index $i = 0, \cdots , N-1$, Node index $j$ of $G$ nodes (or subgroups), step size $\epsilon > 0$, number of iterations $T$} \vspace{0.5em}\\

\hline
 t & & Worker & Communicator\\
\hline

\multirow{7}{*}{0} & 1: &  Randomly draw mini-batch $M^i$ from $X$ & \\
                 & 2: &  Initialize $\Delta w^i = 0$ & Initialize $\Delta w^j = 0$\\
                 & 3: & for all $x \in M$ & \\
                 & 4: & \hspace{5mm} aggregate update $\Delta w^i \leftarrow \partial_w l (w_0 , x)$ &  \\
                 & 5: &  end for & \\
                 & 6: & \textbf{Reduce $\Delta w^i$} to the communicator & \textbf{Reduce $\Delta w^j$} to the communicator and divide by $N$\\
                 & 7: &  & \\
\hline

\multirow{7}{*}{1} & 8: &  Randomly draw mini-batch $M^i$ from $X$ & \textbf{Allreduce} over communicators \\
                 & 9: & \textbf{Broadcast to the workers from the communicator} & \textbf{Broadcast to the workers from the communicator} \\
                 & 10: & update $w_{1} \leftarrow w_0 - \epsilon \Delta w^i$ & \\
                 & 11: &  Initialize $\Delta w^i = 0$ & Initialize $\Delta w^j = 0$\\
                 & 12: & for all $x \in M$ & \\
                 & 13: & \hspace{5mm} aggregate update $\Delta w^i \leftarrow \partial_w l (w_1 , x)$ &  \\
                 & 14: &  end for & \\
                 & 15: & \textbf{Reduce $\Delta w^i$} to the communicator & \textbf{Reduce $\Delta w^j$} to the communicator and divide by $N$\\
                 & 16: &  & \\
\hline

\textbf{\vdots} &  &  \textbf{\vdots}  & \textbf{\vdots} \\

\hline
\end{tabular}
\end{center}
\end{table*}

\subsection{Communication Method}
LSGD divide the jobs of communication and computation (or training) in distributed neural network training, with the former being handled by a CPU and the latter being performed by GPUs, so that the communication burden is not only minimized on GPUs but also hidden by the I/O operations performed by the GPUs.

As shown in Fig. $\ref{LSGD_diag}$, each node is composed of one communicator (triangle) and several workers (circles).
Actual computation (training) is conducted by workers.
The communicators are only responsible for parameter communication such as Allreduce operations.

The conventional distributed SGD (Algorithm 2) needs heavy communication (Allreduce) at every parameter update.
That is, after the computations of gradients of the loss function in each worker, the gradients must be gathered over all workers (line 7 in Algorithm 2). This communication causes more significant bottleneck as more workers are added (see Fig. \ref{fig:CSGD}).

In LSGD, the Allreduce is divided into two layers, local and global.
The workers are involved only in the local Allreduce  communication.
The global Allreduce is performed by communicators and the workers load training data at the same time, as shown in line $8$ of Algorithm 3.
If the data loading time is longer than the Allreduce time over all communicators, then the global Allreduce is hidden and the only non-trivial communication time will be due to the local Allreduce (reduce and broadcasting in line $6$ and line $9$, respectively).

To overlap the global Allreduce and data loading, the parameter update (gradient descent step), which is supposed to take place on line $7$ in Algorithm $3$, is postponed as shown in line $10$ in Algorithm $3$. This overlaps the data loading and the global Allreduce.
For comparison, the parameter update in conventional SGD is completed before going to the next iteration (See line 8 in Algorithm 2).

The flow of the communication can be summarized as follows:
\begin{enumerate}
\item Reduce to the communicator (triangle in Fig. \ref{LSGD_diag}) in each node (grey round rectangle in Fig. \ref{LSGD_diag}).
\item The workers (circles in Fig. \ref{LSGD_diag}) load training data and, at the same time, the communicators conduct Allreduce of all communicators.
\item Broadcast all gathered gradients of parameters from the communicator to the workers in each node.
\end{enumerate}

\subsection{Mathematical Point of View}
As discussed in section \ref{sec:distributed_sgd}, Algorithm $1$ and $2$ are same in the mathematical point of view.
When we have a minibatch $M$ and its partition $\{ M^i \}$ where $i=0, \cdots , N-1$, both algorithms $2$ and $3$ have same parameter values $w_t$ and $w_{t+1}$ just before computing loss function (just before line $4$ in Algorithm $2$ and line $3$ and $12$ in Algorithm $3$).
Therefore, Algorithm $1$, $2$, and $3$ are implement the same SGD formula and we claim that Algorithm $1$, $2$, and $3$ have same accuracy.

\subsection{Summary}
This algorithm splits the communication into intra node communication and inter node communication.
The inter node communication (expensive and slow) is accomplished by only CPUs.
GPUs are involved in only intra node communication (cheap and fast) thereby allowing GPUs to focus on computation rather than communication.
The inter node communication, which is the source of poor scalability, is hidden by the time need for loading data into the GPUs.


Another main advantage of this algorithm is that it doesn't sacrifice accuracy.
Typically, accuracy loss in asynchronous SGD is not avoidable because the latest updates by other workers are delayed.
In contrast, the LSGD algorithm achieves a speed-up by splitting  communication into two layers and reordering necessary operations.

%




%% file: results.tex
\section{Experimental Results}\label{sec:main_results}

To conduct experiment on the proposed algorithm (LSGD), we implemented the LSGD (Algorithm 3) and the CSGD (Algorithm 2) on ResNet-50 (PyTorch implementation \cite{pytorch}) and trained them on ImageNet dataset of ILSVRC2012.

\subsection{Hardware}\label{sec:hardware}


The cluster used in the test is equipped with dual Intel Xeon E5-2695v4 (Broadwell) CPUs (18 cores), dual NVIDIA K80 GPUs (Four Kepler GK210 devices). Each node has SAS-based local storage,  and 256 GB of memory. The nodes are inter-connected with a non-blocking InfiniBand EDR fabric.
 
Each GPU device (Kepler GK210) is assigned to one MPI node (worker node) and one CPU core is assigned for an MPI node for communication. CUDA-aware OpenMPI 3.0.0 is used for MPI communication.
For the largest case, 64 computing nodes (256 GK210 devices) are used with 320 MPI nodes (256 workers and 64 communicators).

\subsection{DNN Framework}

The algorithm is implemented using PyTorch 1.0 \cite{pytorch} with MPI.
For CUDA aware-MPI, OpenMPI 3.0 is linked with PyTorch.

\subsection{Experimental Settings}\label{sec:experimental_setting}

The ImageNet dataset of ILSVRC2012 is used to benchmark the proposed algorithms. Layered SGD (LSGD) and the baseline conventional distributed SGD (CSGD) are implemented based on the PyTorch example code, main.py \cite{pytorch_exmaple}.
ResNet-50 \cite{He_cvpr2016} is the DNN model used in our tests.
Loaded images are normalized by mean = [0.485, 0.456, 0.406] and std = [0.229, 0.224, 0.225].
We use a weight decay level of .0001 and a momentum 0.9.
The learning rate is decreased by $1/10$ per every $30$ epochs.

\subsubsection{Learning rates for Large Minibatches}

The main goal of the experiment is to test the scalability of the algorithm and its accuracy. In our experiments, scaling-up a neural network also leads to larger mini-batch sizes. We set the mini-batch size to be 64 per worker and hence the global mini-batch size is equal to $64 \times N$ where $N$ is the number of  workers.
The benefit of large batch size and methods to preserve accuracy with the large batch size are studied in \cite{You_CoRR2017,imagenetinminutes, imagenet4min,imagenet1hr}.

The learning rate follows the linear scaling rules of \cite{imagenet1hr, imagenet4min}, which keep the ratio of learning rate to minibatch size fixed.
Therefore, as the minibatch size increases, we increase the learning rate linearly. We set one node (four workers) as a base distributed environment. In this base, the whole minibatch size is 256 and the learning rate is $0.1$.
For example, since the learning rate increases linearly as the workers increase, it is $6.4$ when 64 nodes (256 workers) are involved in the experiment.
When we use a large minibatch size and learning rate, 16k and 6.4 in the case of 256 workers, respectively, it accompanies breaking the linear scaling rule and it usually occurs in the early stage of training \cite{imagenet1hr}.
We adopt a warmup strategy \cite{imagenet1hr} to alleviate the large learning rate instability in the early stages of training. This means increasing the learning rate from the base learning rate (0.1 in our case) to the target learning rate (6.4 in our 256 workers case) gradually at every iteration up to a certain epoch (5 epoch in our case).

\subsection{Linear Scalability}

\begin{figure}
\centering
\includegraphics[width=\linewidth]{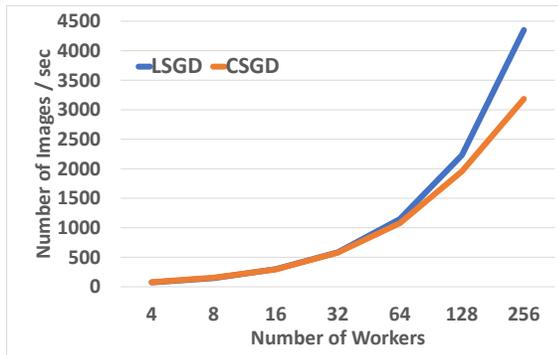}
\caption{Throughput comparison between LSGD and the CSGD.}
\label{fig:throughput}
\end{figure}

\begin{figure}
\centering
\includegraphics[width=\linewidth]{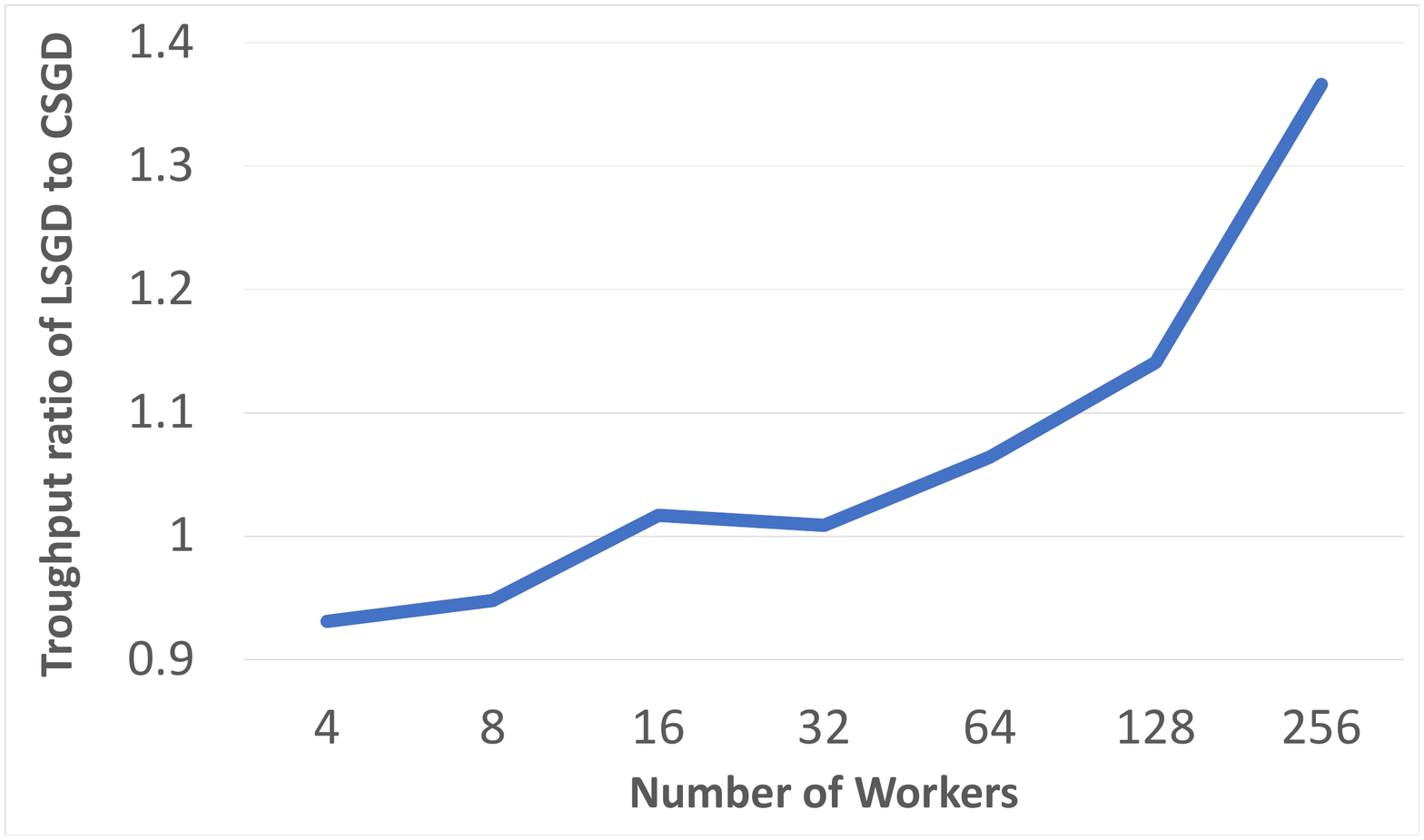}
\caption{The throughput comparison between LSGD and the CSGD. The ratio of throughput of LSGD to the CSGD shown in Fig. \ref{fig:throughput}}
\label{fig:speedup}
\end{figure}

\begin{figure}
\centering
\includegraphics[width=\linewidth]{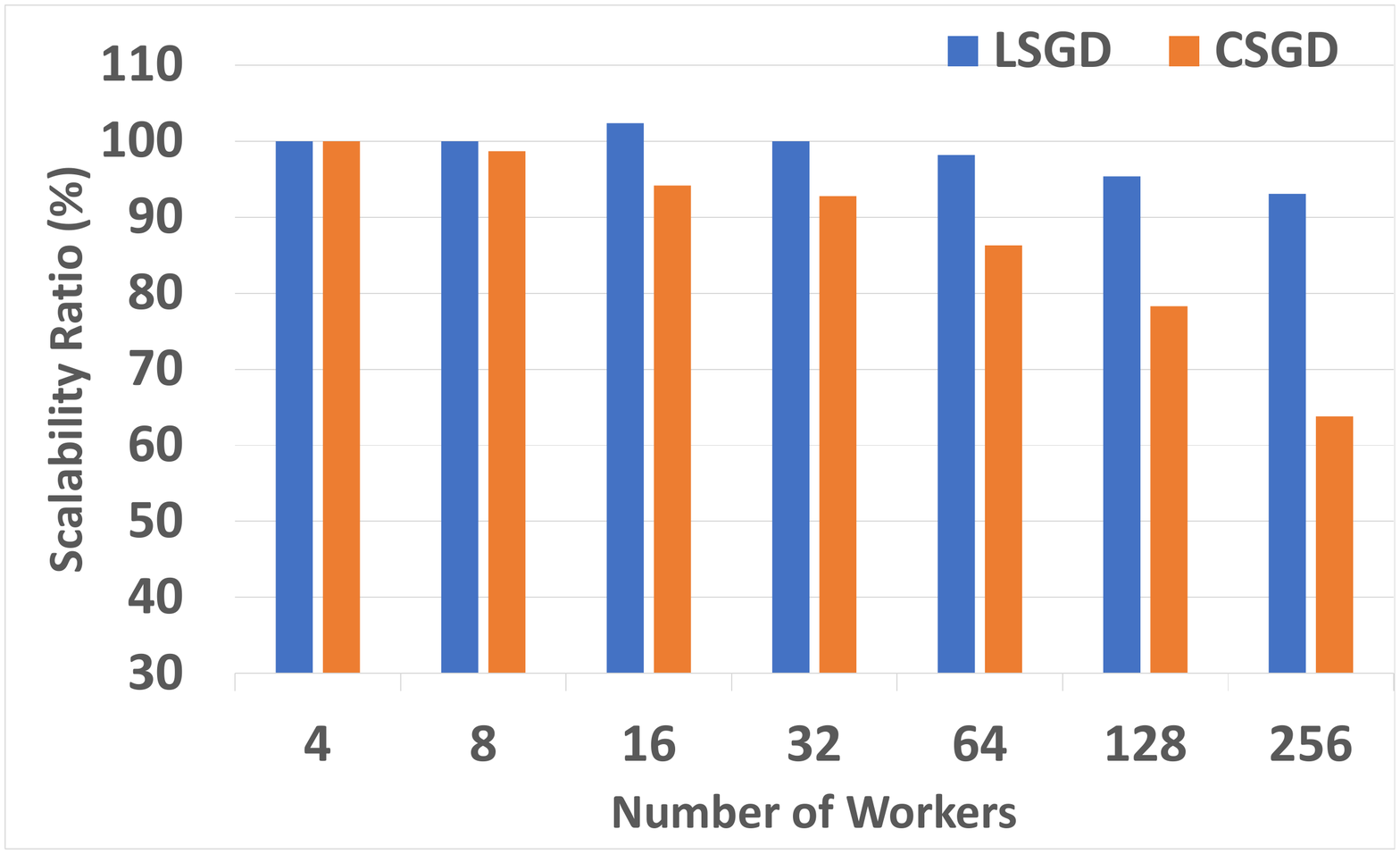}
\caption{Scaling Efficiency. Throughput comparison with perfect linear scalability in percentage.}
\label{fig:scalability}
\end{figure}

A comparison between the training throughput of LSGD and the CSGD is plotted in Fig. \ref{fig:throughput} and Fig. \ref{fig:speedup}.
In Fig. \ref{fig:speedup}, the CSGD is a little bit slower when one or two nodes (four or eight GPUs) are used because of two layer communication.
However, LSGD shows linear throughput as the number of workers increases whereas the throughput of the CSGD decreases as we increase the number of nodes as shown in Fig. \ref{fig:throughput} and Fig. \ref{fig:scalability}.

Figure \ref{fig:scalability} shows the scaling efficiency of LSGD and the CSGD. In this figure, 100 percent means perfect linear scalability. 
In the CSGD, it drops from eight workers (98.7 \%) and continues to drop. It reaches 63.8 \% in the case of 256 works.
On the other hand, the LSGD shows perfect linear scalability up to 32 workers and reaches 93.1 \% in the case of 256 GPUs.

We expect the LSGD will show better linear scalibility when we use bigger data because the LSGD hides the Allreduce time under the data loading time and it can have perfect linear scalibility when the data loading time is longer than the Allreduce time.

\subsection{Accuracy}

\begin{figure}
\centering
\includegraphics[width=0.90\linewidth]{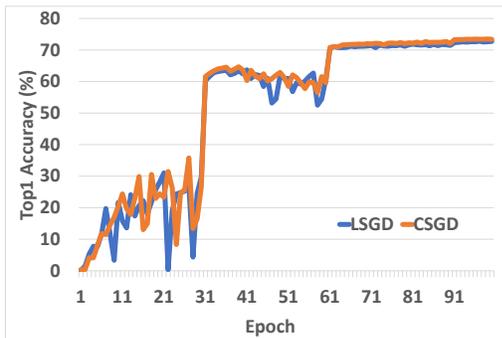}
\caption{Top-1 validation accuracy of LSGD and the CSGD on ImageNet. The number of workers are 256 and the minibatch size 16k. The learning rate is 6.4 and learning rate warmup up to 5 epoch.}
\label{fig:accuracy}
\end{figure}

The accuracy on the ImageNet validation set is plotted in Fig. \ref{fig:accuracy}, for both LSGD and the CSGD algorithms. In the plot, the number or workers is $256$ and the minibatch size is 16,384 (16k).
Both LSGD and the CSGD show similar behavior in this respect.
The best accuracy of LSGD and the CSGD are 72.79 \% and 73.49 \%, respectively.
These accuracies are lower than the accuracy of the original result \cite{he2016deep}.
This accuracy drop results from large minibatch size and the effect of large minibatch size on accuracy is studied in \cite{imagenet1hr} in detail.
The LSGD gradients are unbiased, just as in SGD and the CSGD, and the accuracy curve verifies it.

%% file: conclusion.tex
\section{Conclusion}
In this work we introduced a new variant of SGD for distributed training of deep neural network models, termed Layered SGD. Our efforts were motivated by several limitations of existing distributed training algorithms: Asynchronous SGD is efficient in parameter communication but it suffers from accuracy degradation due to delayed parameter updating, while Synchronous SGD suffers from increasing communication time as the number of compute nodes increases. LSGD takes a divide and conquer approach, and partitions computing resources into subgroups, each consisting of a communication layer and a computation layer. As a result, communication time is overlapped with I/O latency of the workers. 
At the same time, LSGD keeps the formula of the conventional SGD (Algorithm 1). Therefore, LSGD conserve the accuracy of the conventional SGD.

We tested the efficiency of our algorithm by training a deep neural network (Resnet-50) on the ImageNet image classificaton task. Our experiments demonstrated that LSGD has greater throughput compared to the conventional distributed SGD approach. 

In future work, we will investigate the incorporation of LARS \cite{you2017scaling} into our algorithm, as well as comparisons with  other scalable methods \cite{imagenet1hr, imagenet4min, mikami2018imagenet} and deploying LSGD to larger clusters, such as the Summit supercomputer.
Since LSGD is a variabtion of SGD, it is adaptable to any deep neural network, theoretically. So we will apply LSGD to various DNNs to validate its feasibility.